\documentclass[3p,times,procedia]{elsarticle}
\usepackage{ecrc}
\usepackage[bookmarks=false]{hyperref}
    \hypersetup{colorlinks,
      linkcolor=blue,
      citecolor=blue,
      urlcolor=blue}
\usepackage{amsmath}
\usepackage{algorithm}
\usepackage[noend]{algpseudocode}

\usepackage{graphicx}
\usepackage{subcaption}

\usepackage{multirow}

\volume{00}

\firstpage{1}

\journalname{Procedia Computer Science}

\runauth{Vinay A et al.}

\jid{procs}

\usepackage{amssymb}

\usepackage[figuresright]{rotating}

\begin{document}
\begin{frontmatter}



\dochead{8th International Conference on Advances in Computing and Communication (ICACC-2018)}%

\title{Unconstrained Face Recognition using ASURF and Cloud-Forest Classifier optimized with VLAD}


\author[]{Vinay A}
\author[]{Aviral Joshi}
\author[]{Hardik Mahipal Surana}
\author[]{Harsh Garg}
\author[]{K N Balasubramanya Murthy}
\author[]{S Natarajan}

\address{Center for Pattern Recognition and Machine Intelligence, PES University, Bangalore 560085, India}



\begin{abstract}
The paper posits a computationally-efficient algorithm for multi-class facial image classification in which images are constrained with translation, rotation, scale, color, illumination and affine distortion. The proposed method is divided into five main building blocks including Haar-Cascade for face detection, Bilateral Filter for image preprocessing to remove unwanted noise, Affine Speeded-Up Robust Features (ASURF) for keypoint detection and description, Vector of Locally Aggregated Descriptors (VLAD) for feature quantization and Cloud Forest for image classification. The proposed method aims at improving the accuracy and the time taken for face recognition systems. The usage of the Cloud Forest algorithm as a classifier on three benchmark datasets, namely the FACES95, FACES96 and ORL facial datasets, showed promising results. The proposed methodology using Cloud Forest algorithm successfully improves the recognition model by 2-12\% when differentiated against other ensemble techniques like the Random Forest classifier depending upon the dataset used.
\end{abstract}

\begin{keyword}
Face Recognition; Haar Cascade; Bilateral Filter; ASURF; Bag of Words; VLAD; Cloud Forest; Random Forest




\end{keyword}
\cortext[cor1]{Corresponding author. Tel.: 08026721983 ; fax: 08026720886.}
\end{frontmatter}

\email{a.vinay@pes.edu, harshagarwal76@gmail.com}



\section{Introduction}
Face recognition has always been one of the prime topics of research interest for computer vision enthusiasts \cite{s6}. Significant work has been done in the field of face recognition and other affiliated fields for decades. This is because of its application in numerous fields including surveillance systems, authentication systems and other security tools. Such applications can be used for authentication purposes at both public places like theaters, airports etc, and private enterprises for authentication of employees. It can also be used in electronic devices with embedded cameras for authorizing legitimate users. Real-time application of such face recognition models involve the development of fast and efficient models.

This paper proposes one of the efficient models for face recognition which is reliable and computationally inexpensive. To build such a model we have used Haar-Cascade for detection of faces from the image followed by Bilateral filter to eliminate unwanted noise from the detected image. Such a preprocessing step forms one of the important aspects in any image recognition problem. Subsequently, we have used ASURF\cite{asurf1} which makes the model invariant to  translation, rotation, scale, illumination and affine distortion of images. The keypoints detected are characterized using ASURF descriptor which can be further used for classification. The descriptors are quantized using Bag of words\cite{bk6} feature aggregation technique for compact representation. The vector obtained from VLAD and Bag of words model is used for classification using Cloud Forest classifier\cite{c1}.

\section{Related Work}
Face detection is a non-invasive task in the space of object detection. The Haar Cascade technique uses haar-like features improving upon the implementation using the Viola-Jones detector \cite{h1}. According to \cite{h2}, rotated haar-like features were calculated efficiently and improved over the Viola Jones algorithm, thereby reducing the false alarm rate by as much as 12.5\%. As per \cite{h3} which presents an evaluation study using this technique, it is found that in the FA1 and FA2 databases, an accuracy of 100\% is reached while the accuracy for the FEI database is 99.25\% according to the Criterium II benchmark.

Bilateral Filtering has been found to work well in most image processing applications\cite{b1}. It has been used in various contexts such as tone management \cite{b10}, texture editing and relighting \cite{b48}, image noise reduction\cite{b10,b11,b41},  stylization \cite{b72} and demosaicking \cite{b56}. In \cite{b11}, Bilateral Filters are used to enhance low dynamic-range, underexposed videos by varying the exposure in every photoreceptor. Bilateral filters are also used immensely in medical imaging and movie restoration applications. 

Speeded-up robust features (SURF) is one of the well sought after methods for key point detection and description. This method is invariant to in-plane rotation, contrast, scale and brightness. The key point point detector in SURF interpolates the highly discriminative facial points. For the purpose of extracting the physical characteristics from the detected key points, it is sent for description by constructing feature vectors. In order to achieve faster computation time, fast Hessian matrix approximation is used. Also, for the detection of interest points, the scale space is determined by up scaling the integral image based filter sizes. Interest points from the facial images are computed at varied scales where the implementation of scale space is done through an image pyramid. Techniques such as sub-sampling and Gaussian smoothing are incorporated to generate the pyramid labels. For description of interest points the algorithm sets a reproducible orientation through the use of information from the circular area around the derived key point. Subsequently, a square region is constructed according to the chosen orientation. Then descriptors are computed using the key points. The SURF descriptor mainly emphasizes on the spatial distribution of gradient information inside the nearest key point neighborhood.

Although SURF has far-reaching applications, distortions like affine transforms and camera angles in images tend to reduce its accuracy. To overcome the same, a new method termed Affine-SURF (ASURF) \cite{asurf1} was proposed by Yanwei Pang, Wei Li and Yuan Yuan, which overcame these issues. This method was effective in all major applications where SIFT or SURF typically gave poor performance. Affine SURF finds its applications in robust image matching\cite{su1}, automatic identification of cloud cover regions\cite{su2}. 

Vector of Locally Aggregated Descriptors (VLAD) find its use in applications like weakly supervised place recognition\cite{v1}, improvement of image similarity using tensors\cite{v2}, fast video classification\cite{v3}, large scale image retrieval\cite{v4} and event detection\cite{v5}. In \cite{v1}, VLAD is used as one of the important layers of the CNN architecture,i.e., NetVLAD, for image representation which can be used for image retrieval and is readily pluggable as well as amendable to training. In \cite{v3}, VLAD is used in combination with Fisher kernels to outperform the Bag of Words  technique in terms of accuracy. \cite{v4} uses VLAD in large scale image search applications. 

Bag of words find its use in image categorization \cite{bk1}, human action classification\cite{bk2}, multiple-class segmentation\cite{bk3}, medical image retrieval\cite{bk4} and red eye detection\cite{bk5}. \cite{bk3} proposes partitioning of many classes based on objects using bag of keypoints which are combined over mean-shift patches. Also, in \cite{bk2}, a hierarchical model is presented which is characterized as a constellation of bag of words.

\section{Method Proposed}
\makeatletter
\def\BState{\State\hskip-\ALG@thistlm}
\makeatother

An efficient model is proposed with Haar-Cascade for face detection, Bilateral Filters for image preprocessing, ASURF for feature detection and description, Bag of Words and Vector of Locally Aggregated Descriptors (VLAD) for feature aggregation and Cloud Forest for classification.

\subsection{Image detection using Haar-Cascade}


The input image sent to the model is first used for detection of faces from it. For the purpose of face detection from the original image, we have used Haar-Cascade detector. \cite{h1} has the ability of detecting faces rapidly by keeping only information present in grayscale images. The major aspects of Haar-Cascade include an integral image representation of the original image which accounts for quick evaluation of Haar-like features, introduction of an efficient classifier that is modeled using a subset of features from Adaboost and finally a method for integrating more complex classifiers such that they form a cascading architecture. The integral image is computed by sum of pixels above and to the left of the location x,y and given by

\begin{equation}\label{haar_cascade}
ii(x, y) =  \sum_{x' \leq x,y' \leq y} i(x' , y')
\end{equation}

where \(i(x, y)\) is the original image and \(ii(x, y)\) is the integral image.
Finally the detected faces are sent further into the model for image preprocessing and subsequently for classification.

\begin{algorithm}
\caption{Face Detection}\label{euclid}
\begin{algorithmic}[1]
\Procedure{DetectFace}{$image$}\Comment{detects face from input image}
\State $face\gets haar\_cascade.find\_face(image)$
\State \textbf{return} $face$
\EndProcedure
\end{algorithmic}
\end{algorithm}

\subsection{Image preprocessing using Bilateral Filter}

After receiving the detected image, it must be preprocessed to reduce the noise present in it. Noise is defined as random fluctuations in brightness or color information in images. Filtering is generally used for reducing noise in the nearby pixel values which are distant from the signal values.

The method proposed in this paper for filtering uses Bilateral filter, a non-linear, non-iterative and simple technique for blurring an image while taking care of strong edges. \cite{b2} smoothen the image by utilizing a non-linear combination derived by averaging the smooth regions of the nearby image values, preserving edges. Other filtering techniques such as Gaussian blurring \cite{b3} which performs linear operation, neglecting edges are determined by sigma and non-linear filters like median filter \cite{b4}, which replace the pixel values with the median value available in the local neighborhood are outperformed by Bilateral filters. In Bilateral Filtering, weight of the intensity values of the surrounding pixels replace the intensity value of a pixel that is based on Gaussian distribution. This system loops over each pixel and simultaneously adjust weights of the neighboring pixels, retaining sharp edges. Traditional filters generally operate on three separate bands of color resulting in different levels of contrast and smoothing patterns which causes perturbation in the balance of colors and order in which they appear. The method used here uses three bands at once resulting into the average of similar colors and minimizing the artifacts highlighted above. The pixel at value $x$ is replaced with the mean of similar and nearby pixel values. The preprocessed image is sent for keypoint detection further in the model. Bilateral filtering combines the range and spatial domain filters, 
\begin{equation}\label{bilateral_filter}
h(x)  =  \frac{1}{n(x)} \int_{-\infty}^{\infty} \int_{-\infty}^{\infty} g(\xi)c(\xi, x)s(g(\xi), g(x))d\xi
\end{equation}

where n(x) is the normalization factor calculated as, 
\begin{equation}\label{bilateral_filter}
n(x)  =  \frac{1}{n(x)} \int_{-\infty}^{\infty} \int_{-\infty}^{\infty} c(\xi, x)s(g(\xi), g(x))d\xi
\end{equation}

in which c(\(\xi\), x) computes the geometric proximity between a nearby point \(\xi\) and the neighboring center \(x\). s(g(\(\xi\)), g(x)) calculates the photometric similarity between \(x\) and point \(\xi\).

\begin{figure}[h!]
  \centering
  \begin{subfigure}[b]{0.3\linewidth}
    \includegraphics[width=4cm, keepaspectratio]{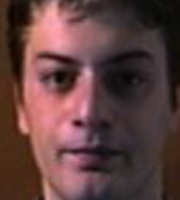}
    \caption{Original Image}
  \end{subfigure}
  \begin{subfigure}[b]{0.3\linewidth}
    \includegraphics[width=4cm, keepaspectratio]{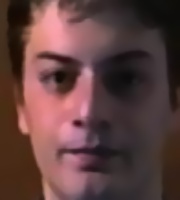}
    \caption{Image after Bilateral Filter}
  \end{subfigure}
  \caption{Reduction in facial noise when bilateral filter is applied}
  \label{fig:bf}
\end{figure}

\subsection{Keypoint detection and description using ASURF}

Smooth apparent deformations are caused in images with smooth boundaries when cameras click these image when placed in varying positions. Affine transforms of an image plane play a good role to locally approximate these deformations. Calculation of affine invariant image local features are the main cause for issues in the domain of object recognition.
Speeded-up robust features (SURF) is invariant to in-plane rotation, contrast, scale and brightness. The key point point detector in SURF interpolates the highly discriminative facial points. For the purpose extracting the physical characteristics from these key points detected it is sent for description of these interest points by constructing feature vectors. In order to achieve faster computation time, fast Hessian matrix approximation is used.

The proposed method performs well when the object has similar illumination conditions and suffers from extreme changes in angle. The method implemented in this paper - Affine-SURF(ASURF) - not only combines the advantages of affine invariance but is also computationally efficient as SURF \cite{asurf1}. The method reliably finds features that have large affine distortions. After this, SURF is applied on all the images. Thus, ASURF effectively overcomes all six mentioned constraints. 

\begin{figure}[h!]
  \centering
  \begin{subfigure}[b]{0.3\linewidth}
    \includegraphics[width=4cm, keepaspectratio]{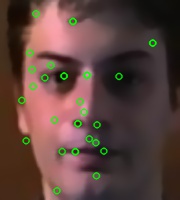}
    \caption{Image with SURF.}
  \end{subfigure}
  \begin{subfigure}[b]{0.3\linewidth}
    \includegraphics[width=4cm, keepaspectratio]{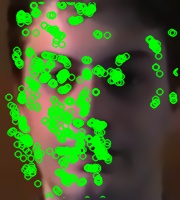}
    \caption{Image with ASURF.}
  \end{subfigure}
  \caption{Change in the number of key points with affine information}
  \label{fig:bf}
\end{figure}

\subsection{ Feature Aggregation using Bag of Words}
After computing the descriptors from ASURF, we have gathered enough information which can be used to distinguish relevant changes in the image parts. For creating the feature vector, a codebook $C$ of clusters is being created. Subsequently each cluster has a set of descriptors which are computed using k-means clustering technique. For every descriptor in an image, it is associated with its nearest centroid of its cluster, $c_{i}$ in the vocabulary. 

Bag-of-words (BoW) model involves counting the number of descriptors in each cluster in the vocabulary. Subsequent steps involve the creation of sparse histogram over the vocabulary, to represent the image in the compact form. Such histograms can be used by classification algorithms to categorize them into different classes. 

\subsection{Feature Aggregation using VLAD}
Similar to Bag of Words (BoW) model, Vector of Locally Aggregated Descriptors (VLAD) is used for representing  images in compact form  based on the locality criteria. In Bag of words a codebook $C$ is constructed of $k$ visual words with the help of k-means. Here, each local descriptor $x$ is related to its closest visual word. In case of VLAD descriptor, the main idea is to accumulate the differences between the descriptor and its associated cluster. The component of $v$ is computed to obtain the sum of all image descriptors which is given by,
\begin{equation}
v_{i,j} = \sum_{x|x=NN(c_{i})} (x_{i}-c_{ij})
\end{equation}

where \(v_{i,j}\) represents the vector over visual words and local descriptor component, x is the descriptor taken for visual word \(c_{i}\). Further, the vector v is L2-normalised by
\begin{equation}
v = \frac{v}{||v||_{2}}
\end{equation}
The quantized vector is sent further for multi-class image classification.

\begin{algorithm}
\caption{Accuracy of Classifier}\label{euclid}
\begin{algorithmic}[1]
\Procedure{GetAccuracy}{$trueLabels,predictedLabels$}\Comment{Calculates the accuracy of the classifier}
\State $CorrectlyClassified\gets 0$
\State $NumberOfLabels\gets 0$

\ForAll{$tl, pl$}  \Comment{iterate over each true(tl) and predicted(pl) label}
\If{tl = pl}
\State $CorrectlyClassified \gets CorrectlyClassified + 1$
\EndIf
\State $NumberOfLabels\gets NumberOfLabels + 1$
\EndFor
\State \textbf{return} $\frac{CorrectlyClassified}{NumberOfLabels}$\Comment{calculate the accuracy}
\EndProcedure
\end{algorithmic}
\end{algorithm}

\begin{algorithm}
\caption{Face Recognition}\label{euclid}
\begin{algorithmic}[1]
\Procedure{FaceRecognition}{$ImagesData,Aggregator, Classifier$}\Comment{Performs face recognition}
\State $ImageDescriptors \gets emptyList()$

\ForAll{$im$}  \Comment{iterate over each image(im)}
\State $face \gets DetectFace(im)$
\State $filteredFace \gets BilateralFilter(face)$
\State $faceKeypoints, faceDescriptor \gets DetectAffineSURF(filteredFace)$
\State $imageDescriptors.append(descriptor)$
\EndFor
\State $vocabulary \gets aggregator.createVocabulary(descriptors)$
\State $NewDescriptors \gets emptyList()$
\State $DescriptorClass \gets emptyList()$
\ForAll{$im, des$}  \Comment{iterate over each image(im) and its corresponding descriptor(des)}
\State $newDescriptor \gets aggregator.computeDescriptor(des, im, vocabulary)$
\State $newDescriptor.append(newDescriptor)$
\State $DescriptorClass.append(newDescriptor)$
\EndFor
\State $trainData, trainLabels, testData, testLabels \gets TrainTestSplit(NewDescriptors, DescriptorClass)$
\State trainingAccuracy = Classifier.train(trainData, trainLabels)
\State predictedOutputs = Classifier.predict(testData)
\State accuracy = GetAccuracy(testLabels, predictedOutputs) \Comment{calculate accuracy}
\EndProcedure
\end{algorithmic}
\end{algorithm}

\subsection{ Classification using Cloud Forest}
CloudForest is a classification algorithm that uses an ensemble of decision trees and is written in Go programming language. Its merit lies in its efficient implementation which fully utilizes the multi-threading potential of modern day machines thus making it fast and flexible. Apart from classification, this method also performs feature selection, regression and structure analysis on heterogeneous data with missing values.

Decision trees are often used for many machine learning  and data mining applications. According to \cite{dt1}, it is unaltering under many feature transformations and is also robust to addition of irrelevant features. In particular, trees that grow deep tend to learn exceedingly irregular patterns, thereby overfitting training datasets. 

Random forests finds its use in all sorts of machine learning applications, ranging from classification to regression. It constructs multiple decision trees with the goal of reducing the variance. It does so by taking the mean of the predictions calculated by the individual trees. This comes at the cost of an increase in bias and loss of interpretability, but greatly boosts the operation of the final model.

\begin{figure}[h!]
  \includegraphics[width=\linewidth]{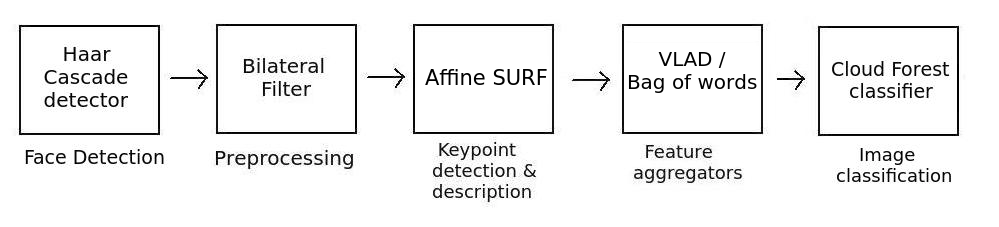}
  \caption{Proposed Approach}
  \label{fig:boat1}
\end{figure}

\section{Performance Analysis}
\subsection{Datasets}
\begin{itemize}
\item FACES95\cite{faces95} contains 1440 images of 72 individuals with 20 images  for each individual being taking using a fixed camera with a delay of 0.5 seconds between successive frames in the sequence. Significant head movement and lighting variations were introduced between images of the same individual. Minor variations in head turn, tilt and expression is present.  
\item FACES96\cite{faces96} is a significantly larger facial dataset which was constructed in a manner similar to FACES95. In total, there are 3040 images for 152 individuals with complex backgrounds. 
\item The ORL\cite{orl_db} Database is relatively small with 40 classes and 400 images. The pictures vary with respect to time and light conditions, facial expression and details.
\end{itemize}

\subsection{Results}

\begin{table}[h]
\caption{Performance of model on multiple datasets and classifiers}
\begin{tabular*}{\hsize}{@{\extracolsep{\fill}}llll@{}}
\toprule
Dataset & Classifier & BoW & VLAD \\
\colrule
\multirow{2}{4em} {FACES95} & Cloud Forest & 97.22\%  & 97.69\% \\
 & Random Forest & 97.45\%  & 95.37\% \\
 & kNN & 64.14\% & 62.36\% \\
 \hline
 \multirow{2}{4em} {FACES96} & Cloud Forest & 94.08\%  & 89.94\% \\
 & Random Forest & 88.07\%  & 77.24\% \\
 & kNN & 61.54\% & 57.21\% \\
 \hline
 \multirow{2}{4em} {ORL} & Cloud Forest & 77.50\%  & 86.67\% \\
 & Random Forest & 84.17\%  & 91.67\% \\
 & kNN & 63.89\% & 71.62\% \\
\botrule
\end{tabular*}
\end{table}


\begin{table}[h]
\caption{Computational efficiency of model on multiple datasets}
\begin{tabular*}{\hsize}{@{\extracolsep{\fill}}llll@{}}
\toprule
Dataset & Classifier & Training Time({\it{in seconds}}) & Testing Time({\it{in seconds}}) \\
\colrule
\multirow{2}{4em} {FACES95} & Cloud Forest & 11.88  & 1.41 \\
 & Random Forest & 139.46  & 0.21 \\
 \hline
\multirow{2}{4em} {FACES96} & Cloud Forest & 103.66  & 5.51 \\
 & Random Forest & 733.33  & 0.60 \\
 \hline
  \multirow{2}{4em} {ORL} & Cloud Forest & 1.75  & 1.22 \\
 & Random Forest & 15.82  & 0.08 \\
\botrule
\end{tabular*}
\end{table}

The model proposed in the paper has been executed on three standard datasets, namely, FACES95, FACES96 and ORL faces. The datasets considered pose variations in head rotation, expression, illumination, scale and affine distortions in the facial images. The model was ran for two different feature aggregators including VLAD and Bag of words with Cloud Forest as classifier. Cloud Forest was further compared with other ensemble classification algorithms like Random Forest and the results are tabulated in Table 1. By setting the classifier as Cloud Forest and running the model with VLAD as the quantization method, it achieves an accuracy of 97.69\%, 89.94\%, 86.67\% on FACES95, FACES96 and ORL respectively in contrast to Bag of words where 97.22\%, 94.08\%, 77.50\% are the accuracies when tested for the same datasets as mentioned before. Also we can conclude that VLAD performs better than Bag of words in most cases. Now upon changing the classifier from Cloud Forest to Random Forest, it can be seen from Table 1 that it could not keep up with the performance in most datasets as that of Cloud Forest except for ORL faces. It gave an accuracy of 95.37\% on FACES95 whereas Cloud Forest proved to be 97.69\% accurate on the same. Similar trend can be seen in case of FACES96 with 77.24\% on Random Forest and 89.94\% on Cloud Forest. The experimentation has been extended for other classifiers like k-nearest neighbours which when tested on the above mentioned datasets like FACES95, FACES96 and ORL achieves an accuracy of 64.14\%, 61.54\%, 63.89\% with Bag of Words as the aggregator and 62.36\%, 57.21\%, 71.62\% with VLAD as the aggregator.

In addition to testing the model on parameters like accuracy, analysis on training and testing time has been performed for deeper insights into the efficiency of the proposed system. On training the model with Cloud Forest as the classifier, it takes much lesser time as compared to Random Forest in all the datasets considered in our case. Such a decrease in time in Cloud Forest can be seen because of the multi-threaded ensemble nature of the classifier. On datasets like FACES95, FACES96, ORL the Cloud Forest algorithm proposed in the paper takes 11.88 seconds, 103.66 seconds, 1.75 seconds to train as compared to 139.46 seconds, 733.33 seconds, 15.82 seconds for the same respectively. Thus, if the application being considered requires training large facial datasets, Cloud Forest classifier is the more appropriate candidate to be considered. However, for time-constrained applications like real time surveillance systems, Random Forest Classifier is more appropriate as testing time is more efficient in this method. The metrics for the time taken for classification is shown in Table 2. 






\clearpage

\normalMode







\end{document}